\documentclass[letterpaper, 10 pt, conference]{ieeeconf}  

\IEEEoverridecommandlockouts                              

\usepackage{tablefootnote}
\usepackage{svg}
\usepackage{multirow} 
\usepackage{censor}
\usepackage{graphicx}
\usepackage{xcolor}
\usepackage{transparent}
\usepackage{import}
\usepackage{textcomp}
\usepackage{array}
\usepackage{float}
\newcolumntype{C}[1]{>{\centering\arraybackslash}p{#1}}
\newcolumntype{L}[1]{>{\raggedright\arraybackslash}p{#1}}
\usepackage{mathtools}
\usepackage[compress]{cleveref}

\title{\LARGE \bf
    Design and Characterization of a Variable-Length \\ Continuum Mechanism with  Force Locking
}

\author{Katelyn King$^{1}$, Veronica Fish$^{1}$, and Allison M. Okamura$^{1}$
\thanks{*This work was supported by a National Science Foundation Fellowship.}
\thanks{$^{1}$Katelyn King, Veronica Fish, and Allison M. Okamura are with the Department of Mechanical Engineering,
        Stanford University, Stanford, CA 94305, USA
        {\tt\small kateking@stanford.edu}}%
}

\begin{document}

\maketitle
\thispagestyle{empty}
\pagestyle{empty}

\begin{abstract}
The utility of flexible continuum mechanisms for dexterous navigation is often impaired by their low stiffness, making them ineffective at manipulation in high-force scenarios. 
To address this challenge, we propose a novel continuum mechanism that achieves both flexible and rigid behavior by antagonistic extension and contraction of a rod-driven continuum helical structure.
The helical design combines variable-length capacity with force locking for workspace and stiffness enhancement.
In this article, we present the detailed design of the proposed mechanism and characterize its performance through experiments that quantify bending and stiffness.
The results demonstrate 180$^{\circ}$ bending range of motion with an average distal positioning error of $<$10\%.
Further tests demonstrate that force locking directly improves axial stiffness and thus indirectly increases bending stiffness anisotropically, with maximum bending stiffness along load paths with a large axial component.
Tensioning the driving rods provides additional stiffness tunability in the force-locked state, where increasing rod tension proportionally increases bending stiffness with a dimensionless gain of 0.56. 

\end{abstract}

\section{Introduction}

Continuum mechanisms, a class of flexible, continuously deforming structures, excel in dexterity and navigation. 
There are a variety of mechanisms that achieve this characteristic bending behavior, including mechanical \cite{dupont2022continuum,feng2021learning,dupont2009design, oliver2021concentric}, fluidic \cite{cianchetti2013stiff}, and magnetic \cite{kim2019ferromagnetic} actuation. 
Compared to fluidic or magnetic alternatives, mechanical systems are simple, responsive, and miniaturizable. 
Mechanically actuated continuum mechanisms consist of a backbone driven by one or more transmission elements (e.g., rod, tube, tendon, wire, spring). 
These kinds of slender structures introduce the fundamental challenge common to continuum mechanisms in general: the trade-off between flexibility and stiffness, where highly flexible continuum mechanisms generally exhibit low stiffness. 
However, applications that leverage continuum robots typically require a range of stiffness, such as low stiffness for safe navigation in delicate environments and high stiffness for large payload manipulation. 

Ideally, one device could perform several tasks with changing or conflicting requirements. 
Recent progress toward such mechanisms includes devices that change their length or stiffness independently \cite{alandoli2024review}\cite{lin2024variable}.
However, few mechanical designs combine variable length \textit{and} variable stiffness; among existing
designs, none can be sufficiently miniaturized due to their complexity. 
Existing mechanical mechanisms use a continuous backbone to vary their length or a discrete backbone to vary their stiffness; these approaches are difficult to unify due to the differences in backbone architecture.
Continuous backbones commonly consist of notched structures produced by selectively removing material from a solid tube; the backbone stretches in the notched regions, enabling length variation via push-pull actuation.
Various patterns of v-shaped \cite{xu2024novel}\cite{zhang2024design} and rectangular \cite{benoist2024tendon} notches have been explored to tune backbone flexibility and extensibility.
Discrete backbones are composed of rigid links \cite{huang2025design} or discs \cite{tamadon2020positioning} chained along axial transmission elements. 
Purely tendon-driven discrete backbones are typically inextensible, since fixed links will not stretch and free links will loosen under compression. 
Instead, tightening the links by increasing the force on the tendons provides a highly effective means of stiffness control, referred to as force locking \cite{chen2024research}.

Force locking relies on contact between backbone elements along the path of the driving transmission element. 
Existing force-locking mechanisms mainly utilize tendon-driven, discrete backbone architectures. 
Tamadon et al. demonstrate stiffness improvement by tensioning a cable-driven manipulator composed of rigid discs that articulate at revolute joints \cite{tamadon2020positioning}. 
Chen et al. present a spine-like continuum robot composed of serial convex-concave joints that they stiffen with a central cable \cite{chen2024research}. 
Kim et al. utilize rolling joints to allow simultaneous articulation and stiffening of rigid links \cite{kim2013stiffness}. 
Discrete links are easily compressed and thus conducive to force locking; however, many continuous backbones preclude this because their notch patterns prevent the gaps from closing fully under tension.
For example, many rectangular notches cannot be fully compressed \cite{benoist2024tendon}, while triangular notches easily close like a hinge \cite{xu2024novel}.
To simplify device design and facilitate small-scale fabrication, a single mechanism is desirable for control of bending, length, and stiffness. 
A continuous architecture is advantageous due to its miniaturizability and extensibility, and augmenting it with force locking presents an attractive method of unifying existing variable-length and variable-stiffness mechanisms. 
To this end, this work investigates the force-locking capacity of a variable-length, rod-driven continuum mechanism with a continuous, helical backbone. 
Unlike localized notches, a continuous helical cut allows the backbone to fully compress, exhibiting the force-locking behavior of free links while preserving a continuous architecture.
We present the detailed design of such a backbone and demonstrate the performance of the mechanism through experiments
that characterize bending and stiffness.

\section{Mechanism Design}

\subsection{Working Principle}

\begin{figure}[t!]
    \vspace{10pt}
    \centering
    \includegraphics[width=0.43\textwidth]{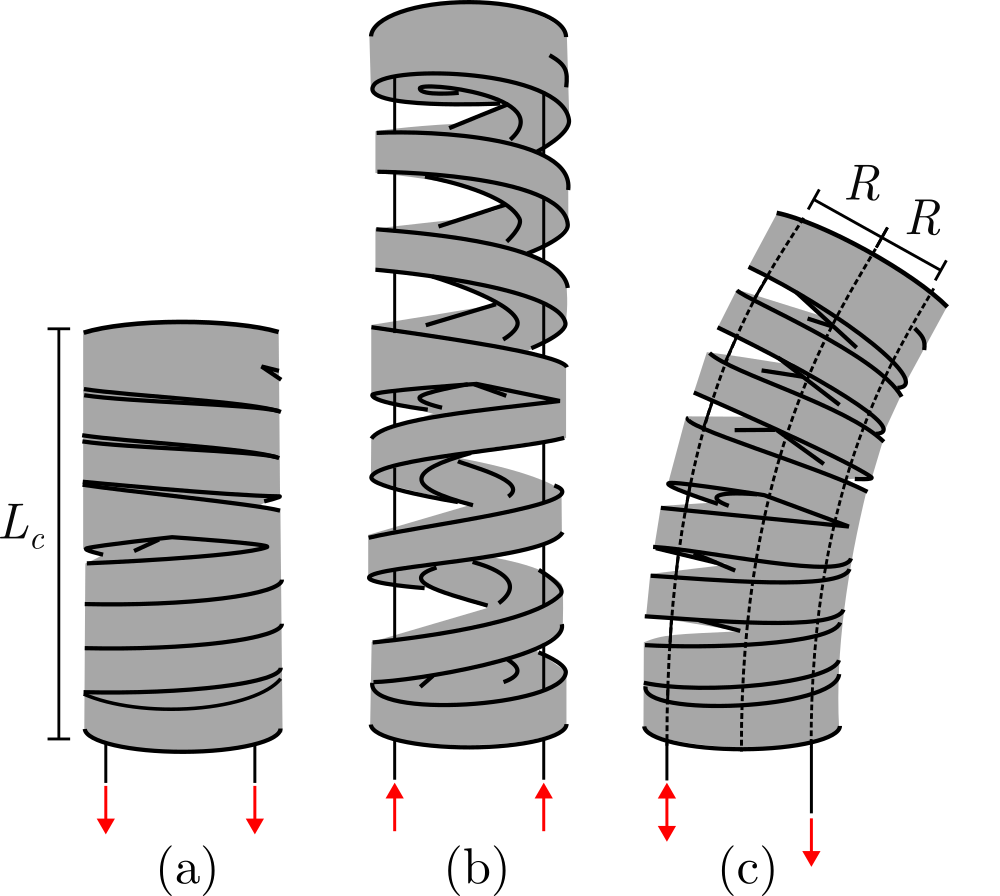}
    \caption{Working principle of the variable-length, force-locking continuum mechanism. (a) Equally pulling both rods contracts the device. The minimum length of the device when it is fully contracted is $L_{c}$. (b) Equally pushing both rods extends the device. (c) Shortening the inner rod relative to the outer rod bends the device.}
    \label{fig:concept}
 \end{figure}

We propose a novel continuum mechanism consisting of a rod-driven, variable-length backbone with a helical architecture (Fig. \ref{fig:concept}). 
We actuate the mechanism with a pair of rods fixed to the distal tip of the backbone and routed through channels positioned $180^{\circ}$ apart around its mean diameter. 
This produces uniform bending bi-directionally by placing the rods equidistant from the centerline, constrains the rods by passing them through channels in the coils, and preserves an open working channel through the center of the device.
It also enables push-pull actuation, which is essential for length control.
The helical geometry of the backbone allows it to stretch; using the rods to push the helical coils apart or pull them together, we extend and contract the backbone, respectively. 
The backbone reaches its minimum length when adjacent coils make contact; this locks the length of the structure at a curvature dependent on its centerline length, producing a force-locking effect.

\subsection{Force Locking}
Force locking is enabled by contact between coils (Fig. \ref{fig:design}); hence, the length and stiffness of the mechanism are coupled, creating distinct flexible and rigid modes.
The flexible mode occurs when there is space between the coils.
Increasing the tension in the inner rod (which follows the inner arc of the device when it is curved) will cause the mechanism to bend rather than stiffen.
The rigid state occurs when the coils make contact along the inner arc of the mechanism.
Increasing the tension in the inner rod does not change the shape of the manipulator, but helps the mechanism resist deformation due to external loads, thereby inducing a stiffening effect.
Importantly, the coupling between length and stiffness does not limit the mechanism's bending range of motion in the rigid state.
Actuating the mechanism with two push-pull rods allows antagonistic extension and contraction, where the outside of the backbone can lengthen even while the inside is contracted.
Thus, varying the insertion length of the outer rod allows the device to bend continuously at any length, including when it is fully contracted in the rigid state.


\subsection{Backbone Architecture}

\begin{figure}[b!]
    \centering
    \includegraphics[width=0.48\textwidth]{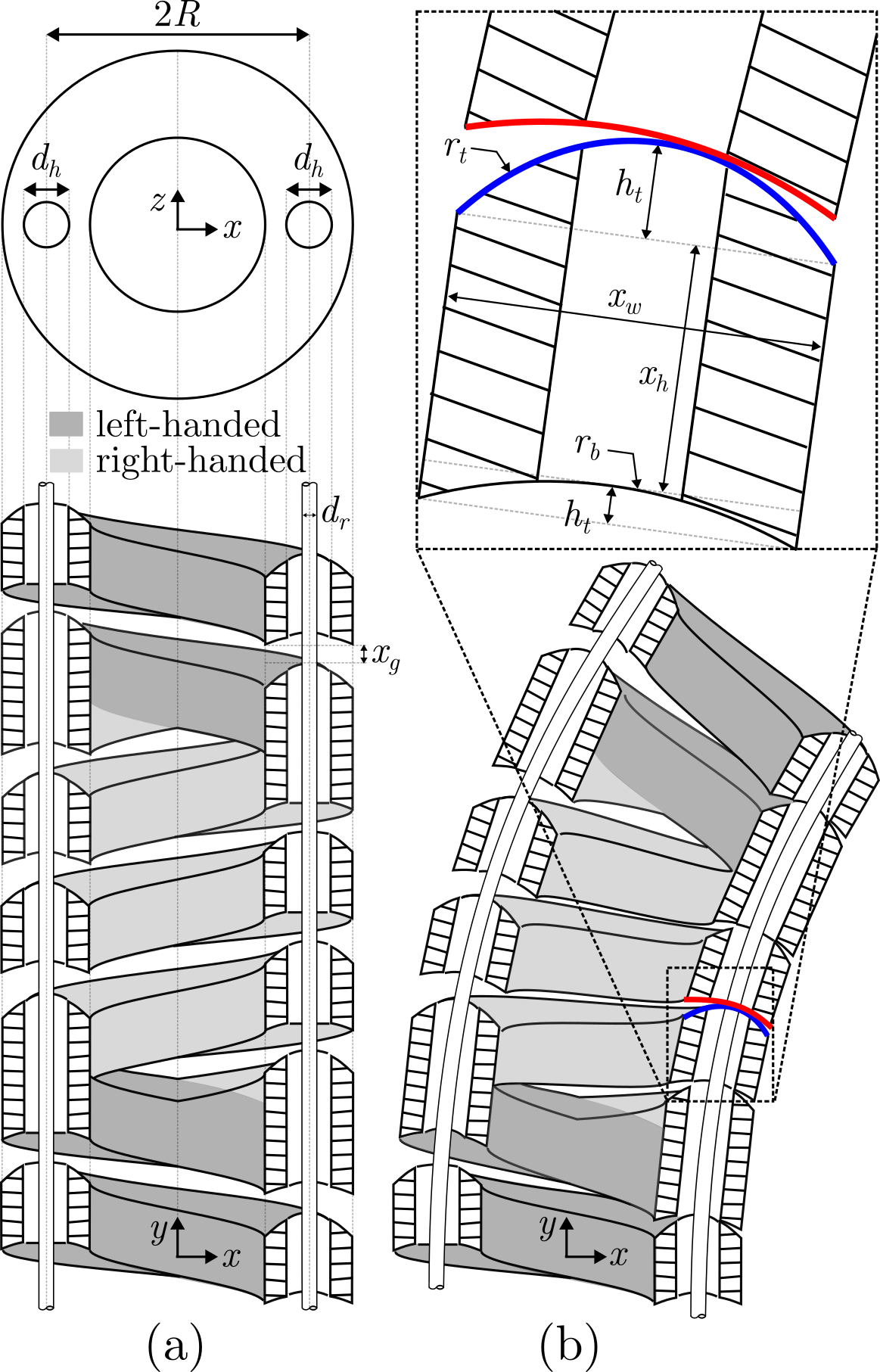}
    \caption{Backbone design showing cross section geometry. As shown, one unit has four coils and adjacent units overlap by half a coil; a complete unit and two partial units are shown. (a) Backbone cross section geometry in the $x$-$z$ plane and in the bending ($x$-$y$) plane. Holes through the center of each cross section facilitate rod-routing. (b) Contact between coils during bending indicating coil cross section geometry.}
    \label{fig:design}
 \end{figure}

The detailed design of the helical backbone is shown in Fig. \ref{fig:design}.
The parameters in Table \ref{tab:params} define the cross sectional geometry and overall dimensions of the device.
The body of the device consists of serial helical unit cells that alternate chirality along the length of the device.
This mitigates out-of-plane deflection during planar bending, as adjacent unit cells are prone to twist in opposite directions. 
Adjacent unit cells are offset by $180^{\circ}$ and overlap by half of a coil, creating a rigid segment between unit cells.
The reversal of chirality between unit cells gives this rigid segment a V-shape, facilitating alignment between active coils when they are compressed and stabilizing the contracted device.
Each helix has a constant pitch and cross section, which is rectangular with a concave top and convex bottom edge, creating a hinge joint between contracted coils in contact. 
Two channels arranged $180^{\circ}$ apart around the mean diameter of the device route the driving rods through the coils. 
Each rod is fixed to the distal tip of the backbone and actuated from its proximal end.
    

\begin{table}[t!]
    \vspace{10pt}
    \renewcommand{\arraystretch}{1.05}
    \centering
    \caption{Geometric Parameters and Nominal Dimensions}
     \label{tab:params}
    \begin{tabular}{C{0.08\textwidth} L{0.22\textwidth} L{0.08\textwidth}}
        \hline
        Symbol & Description & Dimension\\
         \hline
        $x_{w}$ & Coil cross section width & $2$ mm\\
        $x_{h}$ & Coil cross section solid height & $1.29$ mm\\
        $x_{g}$ & Coil gap & $0.5$ mm\\
        $r_{t}$ & Coil cross section top radius & $1.25$ mm\\
        $r_{b}$ & Coil cross section bottom radius & $2.5$ mm\\
        $d_{h}$ & Channel diameter & $0.6$ mm \\
        $d_{r}$ & Rod diameter & $0.5$ mm \\
        $R$ & Mean coil radius & $3$ mm\\
        $T$ & No. coils per unit & $4$\\
        $N$ & No. unit pairs& $3$\\
        \hline
    \end{tabular}
    \vspace{-0.5cm}
\end{table}

\section{Experimental Setup \& Results}

The backbone geometry primarily functions to set the contracted length of the device in the rigid state; however, combinations of parameters that produce equal relaxed/contracted lengths can yield different mechanical properties.
One critical property is axial stiffness, which scales the force required to actuate the mechanism.
The performance goals of the mechanism include low actuation force, a large workspace, and a wide range of bending stiffness.
To this end, the following experiments characterize the axial stiffness, bending range of motion, and bending stiffness of the mechanism.
The effects of backbone geometry, backbone length, and rod forces on the behavior of the mechanism are discussed.

\subsection{Hardware Setup}
\begin{figure}[t]
    \vspace{10pt}
    \centering
    \includegraphics[width=0.48\textwidth]{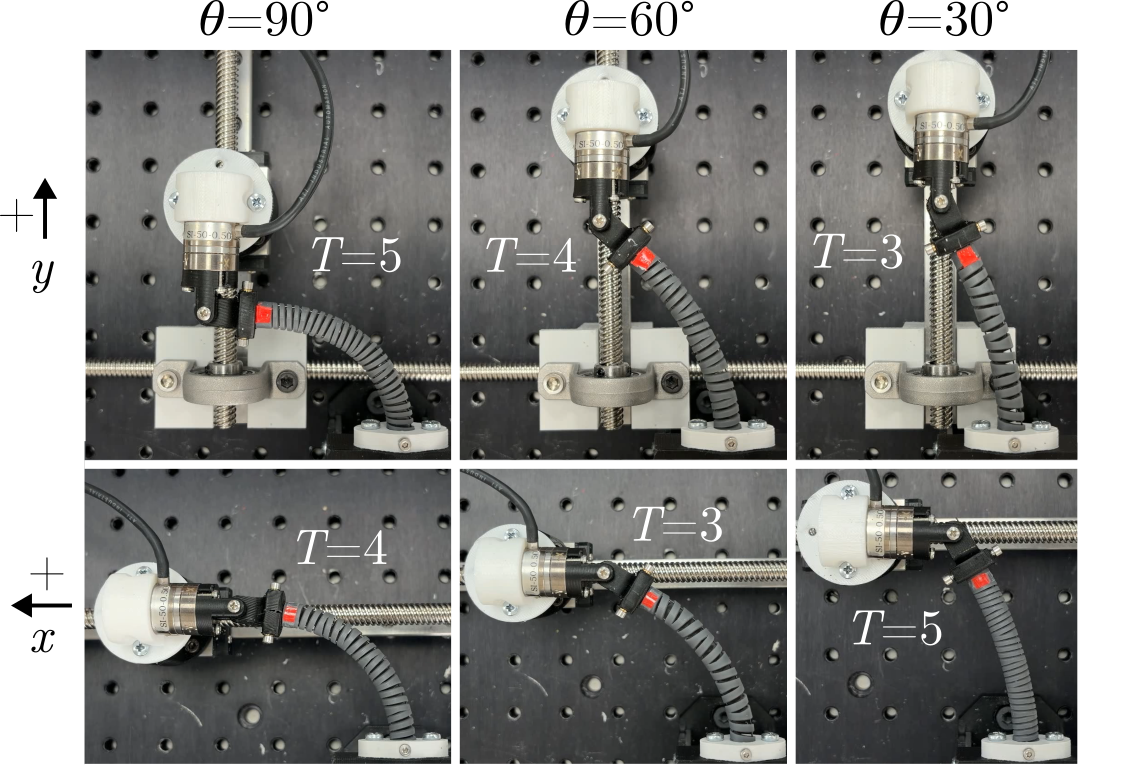}
    \caption{Experimental testbed showing curved poses ($\theta \in [30^{\circ},60^{\circ},90^{\circ}]$) and test samples ($T \in [3,4,5]$). The gantry-mounted force sensor pushes/pulls the tip of each sample $3$ mm along the $x$- and $y$-axes. Stepper-driven lead screws (not pictured) proximal to the fixed base drive the rods.}
    \label{fig:testbed}
    \vspace{-4pt}
 \end{figure}

The experimental testbed consists of two subsystems: (1) an actuation unit to control rod displacement and measure rod force and (2) a two-axis gantry to apply displacement and measure force at the tip of the mechanism.
The actuation unit independently controls each of two superelastic nitinol rods via a stepper-driven lead screw.
A pair of 5-kg load cells (DYLY106, CALT Sensor, Shanghai, China) measures the force applied by the rods.
The gantry utilizes identical stepper-driven lead screws to achieve position control in the $x$-$y$ plane (Fig. \ref{fig:testbed}).
We mount a six-axis force-torque sensor (Nano 17 ATI Industrial Automation, NC, US) on the gantry to displace the tip of the mechanism and measure the applied force for any fixed mechanism pose.
The custom components for both subsystems are 3D-printed from polylactic acid (PLA) filament with a Bambu X1 printer.

\subsection{Prototype Fabrication}
\label{sec:prototypes}

\begin{figure}[t]
    \centering
    \includegraphics[width=0.4\textwidth]{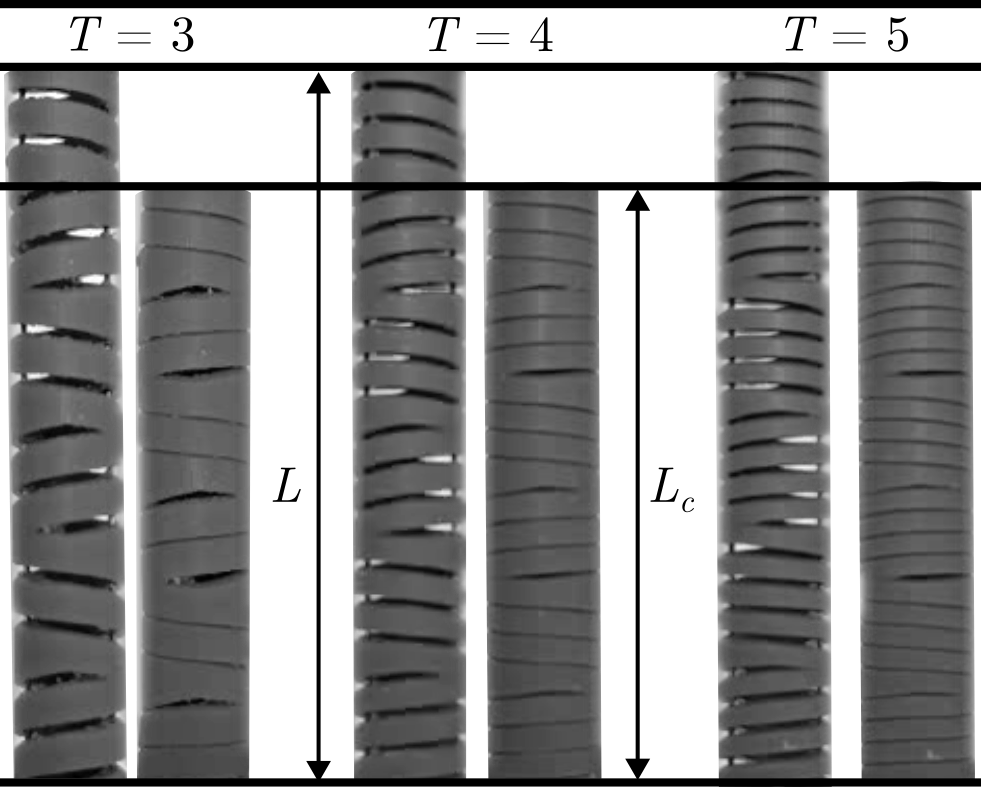}
    \caption{Manufactured samples showing differences in coil geometry. Small variations in contracted length were observed, but are not visible at this scale.}
    \label{fig:samples}
 \end{figure}

To characterize the effect of backbone geometry on the behavior of the mechanism, we fabricated three prototypes using a Form3+ SLA printer and Tough2000 Resin v1 ($E \approx 2200$ MPa) (Formlabs, Somerville, MA, USA), which qualitatively demonstrates the desired combination of strength and flexibility (Fig.~\ref{fig:samples}). 
Each prototype has a unique number of coils per unit ($T \in [3,4,5]$); we also adjusted the solid height of the coils ($x_{h}$) and coil gap ($x_{h}$) to maintain a constant contracted and relaxed length across the three samples (Table \ref{tab:sample-params}).
The remaining parameters defined in Table \ref{tab:params} remain constant across the three samples.

\begin{table}[t!]
    \vspace{5pt}
    \renewcommand{\arraystretch}{1.05}
    \centering
    \caption{Geometric Dimensions of Test Samples}
    \label{tab:sample-params}
    \begin{tabular}{C{0.03\textwidth} C{0.11\textwidth}  C{0.11\textwidth} C{0.11\textwidth}}
        \hline
        & $T=3$ & $T=4$ & $T=5$ \\
         \hline
        $x_{h}$ & $1.98$ mm & $1.29$ mm & $0.90$ mm\\
        $x_{g}$ & $0.81$ mm & $0.5$ mm & $0.33$ mm\\
        \hline
    \end{tabular}
    \vspace{0.1cm}
    \\ \footnotesize \emph{{See Table~\ref{tab:params} for variable definitions.}}
    \vspace{-0.5cm}
\end{table}

\begin{figure*}[t]
   \vspace{10pt}
    \centering
    \includegraphics[width=.98\textwidth]{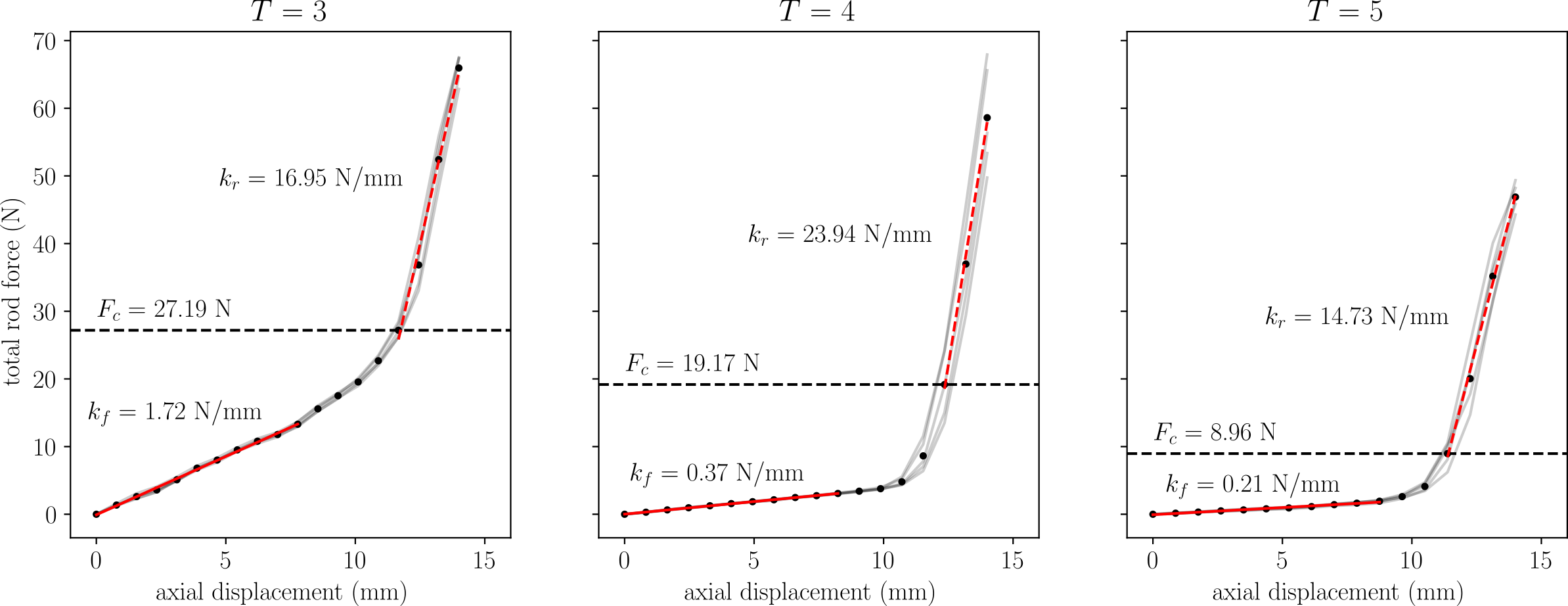}
    \caption{Force-displacement curves quantifying the axial stiffness of three backbone prototypes with varied coils per unit. Gray lines represent the sum of the tension applied by the inner and outer rod while pulling the backbone from its relaxed length to its fully contracted length in each of the five trials. The black points are the average of all five trials. Compressing the backbone first deforms the helical segments which behave like a Hookean spring with stiffness $k_{f}$ indicated by the slope of the solid red line. Once the coils are fully compressed, the flexures begin to close, introducing non-linearity in the curve. The horizontal line estimates the minimum force ($F_{c}$) required to fully contract each backbone. After this point, all the notches are closed, requiring more force to further compress the material, quantified by the slope of the dashed red line ($k_{r}$).} 
    \label{fig:axial-test}
 \end{figure*}

To estimate the relaxed length of the backbone, we take the product of the unit count, coils per unit, and helical pitch,
\begin{equation}
L = 2N\left(T-\frac{1}{2}\right)(x_{h}+h_{t}+h_{b}+x_{g}),
\end{equation}
where $h_{t,b}$ are the depths of the convex/concave regions of the coil cross section (Fig. \ref{fig:design}b),
\begin{equation}
h_{t,b} = r_{t,b} - \sqrt{r_{t,b}^{2}-\frac{1}{4}x_{w}^{2}}.
\end{equation}
To calculate the contracted length of the backbone, we assume (1) the coil gaps and (2) flexures between helical units contract fully. 
Subtracting the length of these notches from the relaxed length of the coil yields the contracted length, which simplifies as follows:
\begin{equation}
L_{c} = \left(2N\left(T -\frac{1}{2}\right) -\frac{1}{2}\right)(h_{t}+x_{h}) + \left(N-\frac{1}{2}\right)(h_{b}+x_{g}).
\end{equation}

We estimate a contracted length of $38.5$ mm and relaxed length of $52.5$ mm for the samples in Table \ref{tab:sample-params}.
After measuring the manufactured samples, we found the relaxed length was consistent across samples while the contracted length varied up to 10\% from the prediction.
The model for $L_{c}$ underestimates the true contracted length, likely due to the assumption of complete closure of the flexures between units.
Qualitatively, we observed that samples with larger $x_{h}$ did not fully contract in the flexures between segments (Fig.~\ref{fig:samples}); we attribute this to increased axial stiffness.

\subsection{Axial Stiffness}

To characterize variation in axial stiffness across the three samples described in Section~\ref{sec:prototypes}, we mounted each prototype on the actuation unit (Fig. \ref{fig:testbed}), pulled the rods to compress each sample from its relaxed to contracted length, and measured the resulting rod forces.
The force-displacement curves obtained from five repeated trials are plotted in Figure \ref{fig:axial-test}.
 
\subsection{Bending Range of Motion}

\begin{figure}[t!]
    \vspace{5pt}
    \centering
    \includegraphics[width=0.46\textwidth]{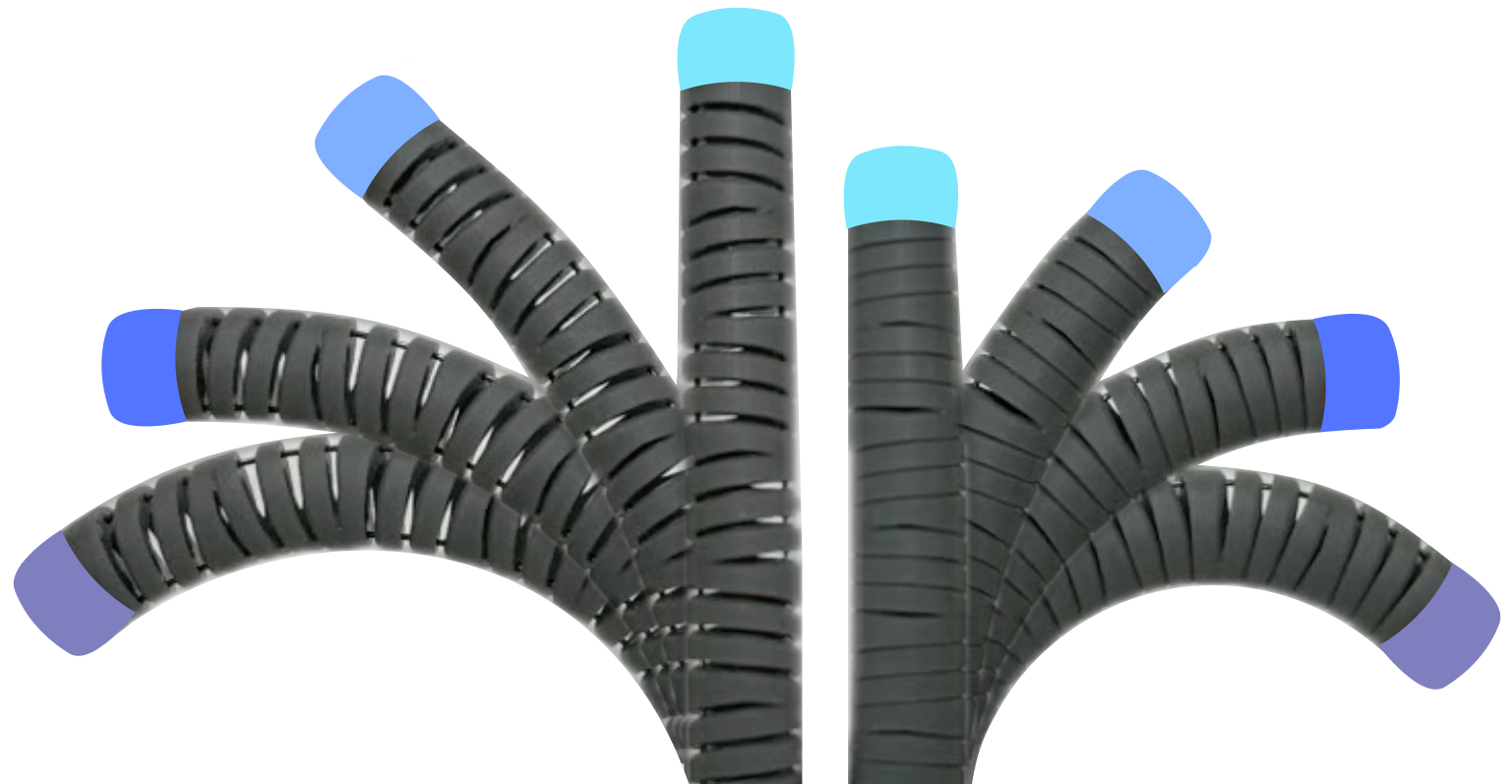}
    \caption{Relaxed (flexible) v. contracted (rigid) backbone ($T = 4$) at various bending curvatures. In the rigid state, at least one rod is fully contracted in each case to achieve the force-locked state.}
    \label{fig:colorful}
    \vspace{-4pt}
 \end{figure}

\begin{figure}[t]
    \vspace{10pt}
    \centering
    \includegraphics[width=0.4\textwidth]{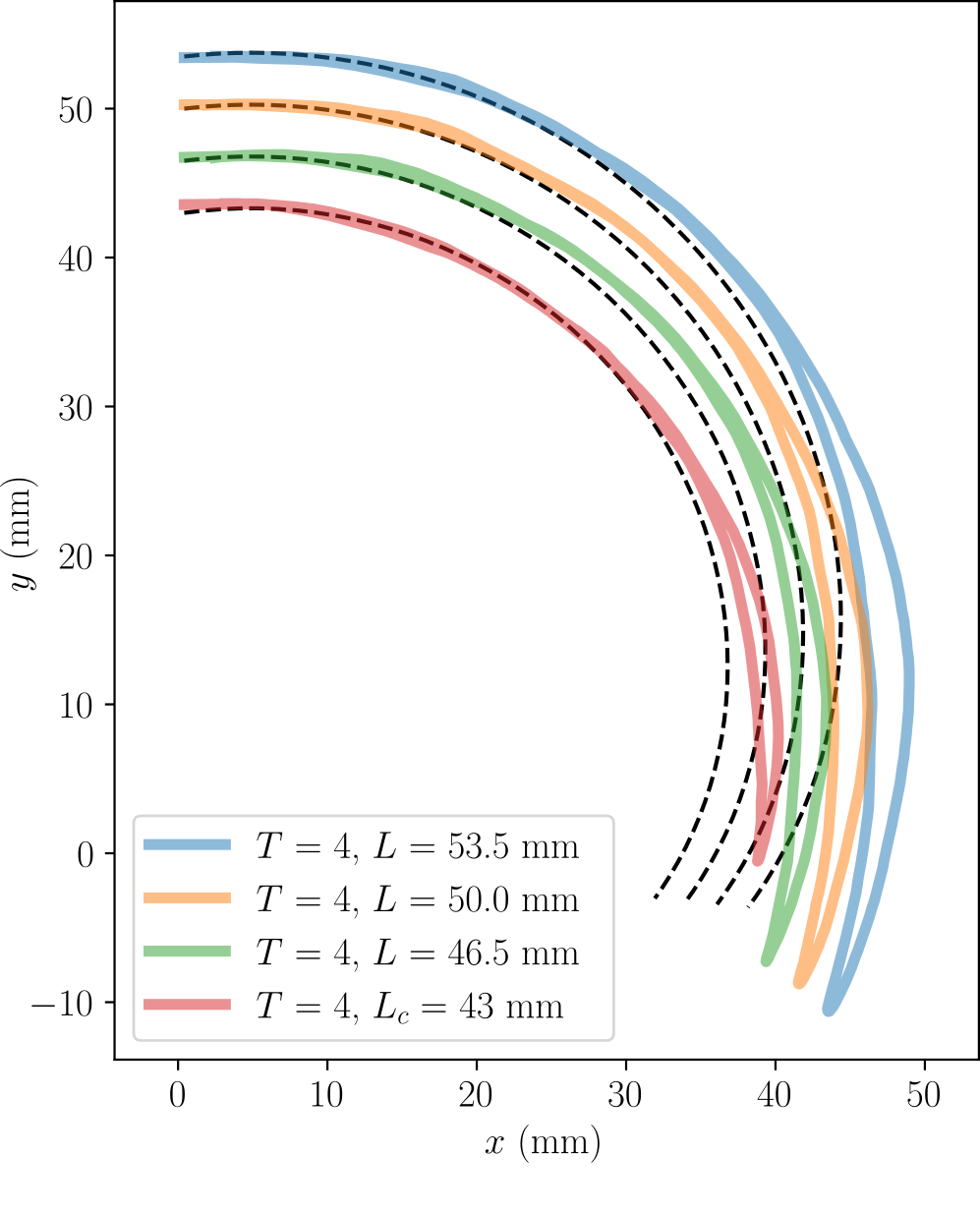}
    \caption{Bending curvature in the $x$-$y$ plane. Pushing and pulling the outer rod bends the mechanism in the range of [$0^{\circ}$, $180^{\circ}$]. The fixed length ($L$) of the inner rod (equal to the uncurved starting length of the backbone) is indicated for each trajectory. The fully contracted length ($L_{c}$) maps the rigid workspace, while all other lengths are within the flexible workspace. The dashed lines indicate the theoretical constant curvature bending trajectory. High friction results in large hysteresis error at high curvatures.}
    \label{fig:bendingRoM}
\end{figure}

A visualization of the bending workspace in a set of relaxed and fully contracted configurations is provided in Fig.~\ref{fig:colorful}. To characterize the bending range of motion of each of the three prototypes described in Section~\ref{sec:prototypes} in the $x$-$y$ plane, we drove each device through bending angles in the range of [$0^{\circ}$, $180^{\circ}$] for four uncurved starting lengths ranging from fully contracted to relaxed.
To actuate the mechanism, we pushed/pulled the outer rod only, keeping the inner rod in tension at a fixed length. A video of each trial was processed with OpenCV in Python to measure the distal tip position.
The experimental trajectories obtained from three repeated experiments and the theoretical values based on the constant curvature assumption are plotted in Figure \ref{fig:bendingRoM}.
Due to hysteresis, positioning error increased at higher curvatures; Table \ref{tab:bending-error} quantifies hysteresis error along with average deviations throughout the trajectory. 

\begin{table}[t]
    \vspace{5pt}
    \renewcommand{\arraystretch}{1.05}
    \centering
    \caption{Positioning and Hysteresis Error}
    \label{tab:bending-error}
    \begin{tabular}{C{0.015\textwidth} C{0.015\textwidth} C{0.08\textwidth}  C{0.08\textwidth} C{0.08\textwidth} C{0.08\textwidth}}
        \hline
         $T$ & & Average position error (mm) & Maximum position error (mm) & Average hysteresis error (mm) & Maximum hysteresis error (mm) \\
         \hline
        $3$ & $L$  & $3.12$  & $9.62$  & $1.44$  & $3.66$ \\
              & $L_{c}$  & $5.10$  & $9.13$  & $0.99$  & $3.15$ \\
        $4$ & $L$ & $3.22$  & $7.91$  & $2.72$  & $6.48$ \\
              & $L_{c}$ & $5.60$  & $7.86$  & $1.12$  & $2.73$ \\
        $5$ & $L$ & $3.99$  & $7.16$  & $3.57$  & $7.01$ \\
              & $L_{c}$ & $3.05$  & $6.56$  & $1.38$  & $3.68$ \\
        \hline
    \end{tabular}
\end{table}


\subsection{Bending Stiffness}
 
\begin{figure*}[t]
    \vspace{10pt}
    \centering
    \includegraphics[width=\textwidth]{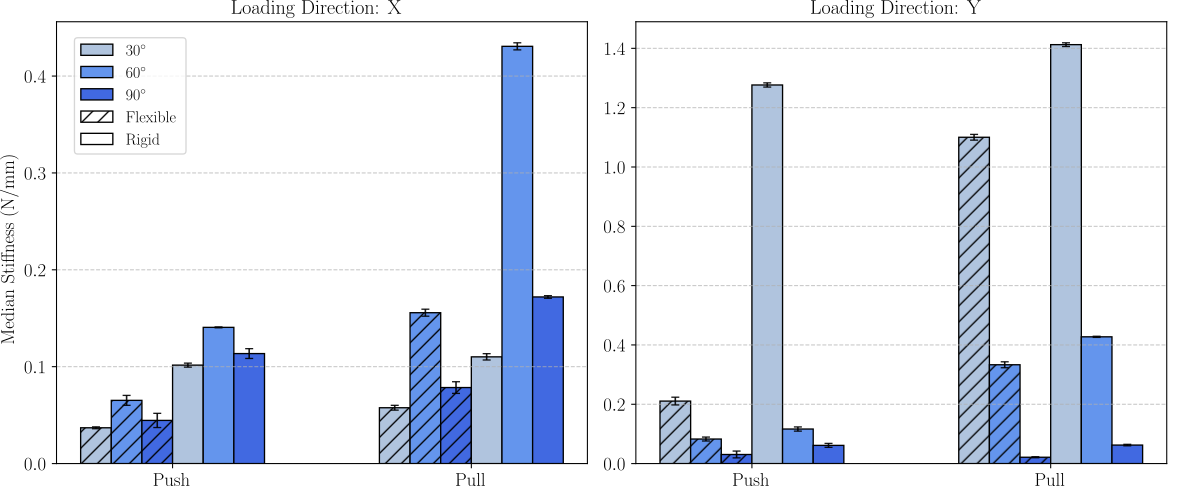}
    \caption{Bending stiffness in each curved pose ($30^{\circ},60^{\circ},90^{\circ}$) in each loading direction ($\pm x,\pm y$) for backbone geometry with $T=4$. Stiffness is higher in the rigid state compared to the flexible state in all conditions. Stiffness is anisotropic, with higher forces measured along the $y$-direction than the $x$-direction. In the $x$-direction, pulling stiffness is generally higher than pushing. In the $y$-direction, stiffness decreases with increasing bending angle, consistent with decreasing axial load. }
    \label{fig:stiffness-test}
 \end{figure*}

\begin{figure}[h!]
\centering
\includegraphics[width=0.48\textwidth]{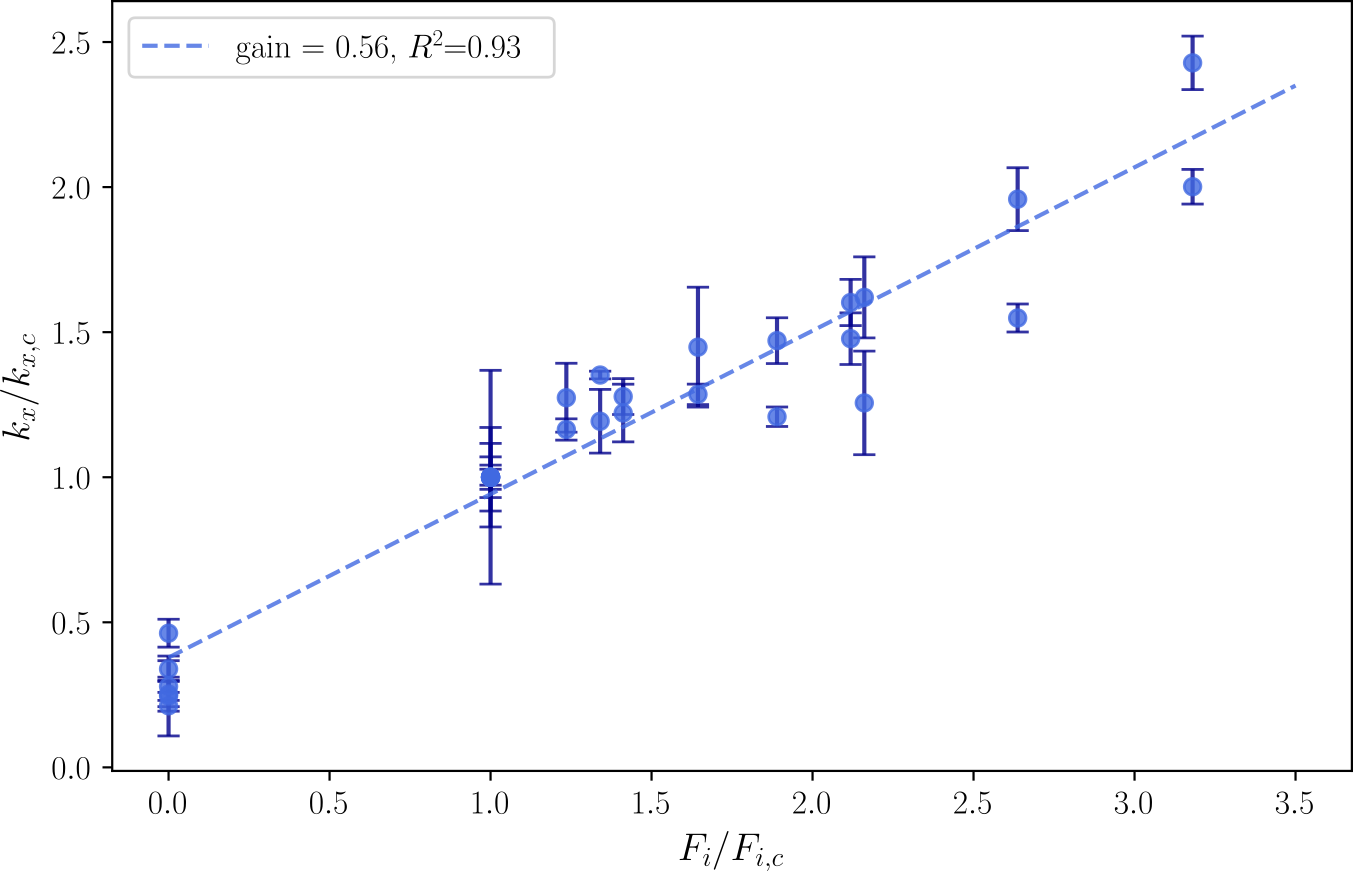}
\caption{Effect of increasing rod tension on the stiffness of the uncurved ($0^{\circ}$) mechanism loaded along the $x$-axis. Tension is normalized relative to the minimum inner rod tension required to fully contract the backbone ($F_{i,c}$); in the uncurved state, equal tension is applied to the inner and outer rods. $k_{x,c}$ is the stiffness measured for $F_{i,c}$. Data are aggregated across backbone geometry and push/pull loading directions, as all conditions exhibit the same positive trend.}
\label{fig:rod-force-test}
\vspace{-10pt}
\end{figure}

To characterize the bending stiffness of the mechanism, we applied a fixed displacement to the tip of the device and measured the resulting force.
For each sample, we tested four poses corresponding to bending angles of $0^{\circ}, 30^{\circ}, 60^{\circ},$ and $90^{\circ}$ along the $-x$-axis.
For each pose, we tested the mechanism in its rigid and flexible states. 
In the flexible state, the backbone was at its relaxed length and no force was applied to either rod.
In the rigid state, we pulled the inner rod to its fully contracted length and conducted four trials with increasing tension applied to the inner rod.
We defined the baseline rigid tension as the force exerted by the inner rod upon initial contact between the coils at the desired pose, and incremented the tension by 50\% of the baseline over the next three trials. 
In each trial, we applied five cycles of push-pull displacement to the tip of the mechanism using the gantry-mounted force sensor.
One cycle consisted of pushing the tip 3 mm along the $+x$-axis, then pulling 6 mm along the $-x$-axis, and finally pushing 3 mm along the $+x$-axis.
We repeated the same procedure to estimate stiffness along the $y$-axis at $30^{\circ}, 60^{\circ},$ and $90^{\circ}$.
We omitted $0^{\circ}$ along the $y$-axis, as loading the rigid device axially would damage the force sensor.
To demonstrate the effect of contracting the backbone, Figure \ref{fig:stiffness-test} compares bending stiffness in the flexible and rigid states.
To quantify the effect of rod force, Figure \ref{fig:rod-force-test} plots the change in stiffness obtained by increasing inner rod tension in the rigid state. 

\section{Discussion}

We hypothesize that the design of the backbone, applied rod force, and force locking are the main factors affecting mechanism performance; the following discussion considers the effect of each on axial stiffness, bending stiffness, and bending curvature. 

\subsection{Effects of Force Locking}
Contraction is the primary mechanism by which the backbone varies its stiffness; as the backbone shortens, the observed stiffness gain is a coupled effect of both force locking and the reduced moment arm of the shortened backbone.
Figure \ref{fig:axial-test} indicates a sharp increase in axial stiffness at the inflection point where the backbone fully contracts (Fig.~\ref{fig:axial-test}).
As a result, bending stiffness in the rigid state is anisotropic, with the highest stiffness occurring in loading directions with an axial component. 
Across all poses, we measured higher stiffness along the $y$-axis than the $x$-axis, showing that the increase in axial stiffness due to coil locking also contributes to bending stiffness.
Additionally, we measured higher stiffness when pulling along the $x$-axis compared to pushing in curved poses ($30^{\circ},60^{\circ},$ $90^{\circ}$).
This is explained by coil locking, since pulling with the curvature of the device further compresses the backbone.
When the mechanism is uncurved, both sides of the backbone are contracted, so pushing compresses the far side of the backbone in the same way pulling compresses the near side; accordingly, we  measured similar stiffness when pushing and pulling the  along the $x$-axis with the backbone uncurved ($0^{\circ}$).

\subsection{Effect of Rod Forces}
In the same way that external forces can further compress the backbone, higher actuation forces more tightly compress the coils, contributing to axial and bending stiffness.
Figure \ref{fig:rod-force-test} shows that increasing the tension in the rods improves the stiffness of the mechanism in the uncurved state.
Higher tension magnifies force locking by tightly compressing the coils. 
Increasing rod force also pretensions the backbone, which can counteract moments from external loads.
When the backbone is curved, the force applied by the inner rod is the primary mechanism that resists pushing deformation along the $x$-axis, since the coils along the outer rod are not fully compressed.
To resist pulling deformation along the $x$-axis, the mechanism relies on force from the outer rod and force locking between coils along the inner rod; however, the tension in the outer rod must be less than the inner rod to curve the backbone, so force locking is useful to help resist pulling forces along the $x$-axis.


\subsection{Effect of Backbone Geometry}
The backbone geometry primarily functions as a geometric constraint that sets the rigid length of the backbone; however, the geometry should be tuned to achieve acceptable axial and bending stiffness.
Minimizing the spring rate of the helical units will increase the variable range of bending stiffness the backbone exhibits; increasing the number of coils per unit ($T$) will greatly reduce axial stiffness (Figure \ref{fig:axial-test}).
Additionally, more coils per unit will increase the notches in the backbone, allowing the coil to better approximate a continuous arc and helping to reduce positioning error.
The results in Table \ref{tab:bending-error} show the maximum positioning error increases as $T$ decreases; this is due to a combination of fewer notches impeding smooth bending articulation and increased axial stiffness resulting in higher rod forces/friction.
In our experiments, the stiffer prototype ($T = 3$) primarily deformed at the base, indicating high friction limiting the tension transferred to the tip.

\section{Conclusions and Future Work}

We present a novel variable-length, force-locking  mechanism based on a rod-driven continuum helical structure that achieves both flexible and rigid behavior.
The device demonstrates the efficacy of force locking applied to a continuous backbone structure, unifying variable length and variable stiffness capacity in a miniaturizable device.
The primary goal in designing the backbone geometry is to set the minimum length of the device in the rigid state; coil parameters should be selected accordingly, but must then be tuned to achieve the desired axial stiffness. 
Low axial stiffness is desirable because it reduces the necessary force applied by the rods to achieve bending and contraction when the device is in its flexible state.
In the device's rigid state, coil compression increases axial stiffness, anisotropically improving bending stiffness in loading directions with an axial component.
This approach simplifies control and small-scale fabrication compared to other continuum mechanisms that use decoupled methods for length and/or stiffness variation.

While desirable for miniaturization, the simplicity of this device presents some limitations.
Due to the coupling between curvature and stiffness control, the rigid workspace is small compared to the flexible range of motion, preventing the device from achieving high stiffness throughout its workspace.
Additionally, the device is currently limited to a single plane of bending, while many other mechanisms in the literature have additional degrees of freedom \cite{luo2023novel,iqbal2025continuum}.

Thus, future work will investigate complex bending behavior beyond planar motion, including 3D navigation and multiple curvatures.
Currently, the backbone is asymmetric at the flexures between helical units, causing the contracted length to vary slightly around the circumference of the backbone.
We hypothesize that alternating the chirality of adjacent segments improves torsional rigidity, but need to quantify the significance of this compared to a simpler, continuous helical backbone.
This would yield uniform bending behavior omni-directionally, enabling the addition of a third and/or fourth rod to realize 3D curvatures.
With additional transmission elements, it is possible to lock a larger section of the backbone by contracting multiple rods; future work will investigate the three-dimensional stiffness of the mechanism to determine whether multiple contracting rods can improve the stiffness and mitigate the stiffness anisotropy observed in the planar mechanism.
Additional changes to the design of the backbone may include the coil geometry; the question remains as to whether the convex/concave cross section provides more stable locking and improved stiffness compared to a simple, rectangular cross section.

Finally, we seek to reduce friction in the mechanism to mitigate hysteresis in bending; changes to the backbone geometry would allow us to explore fabrication methods beyond 3D printing, opening a greater selection of low-friction materials. 
In the short term, we will explore low-friction sheaths to allow the rods to slide more smoothing in the routing channels. 

Ultimately, we will explore applications that best leverage the combined benefits of variable length and variable stiffness.
A key application area for continuum robots is minimally invasive surgery, which requires tools that can both navigate tortuous pathways in the body and transmit sufficient force to manipulate tissue. For example, in gastrointestinal endoscopy, navigating the hollow GI tract and applying sufficient force to biopsy tissue requires flexible position control along the endoscope and high force application at the target site \cite{kim2024robot}; our mechanism may be applied as an endoscopic overtube to facilitate these goals via its flexible motion and force-locking capacity. 
Depending on the location of lesions within the GI-tract, an endoscopic end-effector may need to bend up to $180^{\circ}$, which is easily achieved with our mechanism. 
An additional application is bronchoalveolar lavage, a minimally invasive procedure that requires navigating tortuous lung anatomy to diagnose pathologies in the distal branches of the bronchoalveolar tree \cite{rothe2024model}.
In the flexible, low-stiffness state, our mechanism can facilitate navigation in these delicate pathways, and control the force applied when biopsying tissue.
Overall, our mechanism provides a wide range of curvature, length, and stiffness control, making it a useful tool for tasks with changing or conflicting requirements such as a minimally invasive surgical procedure.











\bibliographystyle{IEEEtran}
\bibliography{references}

\end{document}